\documentclass[conference]{IEEEtran}
\usepackage[utf8]{inputenc}
\IEEEoverridecommandlockouts
\usepackage{array}
\usepackage{graphicx}
\usepackage{subcaption}
\usepackage{placeins}
\usepackage{enumitem}
\usepackage[utf8]{inputenc}
\usepackage{amsmath,amssymb,amsfonts}
\usepackage{multirow}
\usepackage{graphicx}
\usepackage{cite}
\usepackage{url}
\usepackage[font=small]{caption}
\usepackage{algorithm}
\usepackage{algorithmic}
\usepackage{hyperref}
\usepackage{booktabs}
 
\begin{document}
\newcommand{\sysName}{QFed}
\title{{\sysName}: Parameter-Compact Quantum-Classical Federated Learning}
\author{\IEEEauthorblockN{ Samar Abdelghani, Soumaya Cherkaoui}\\
\IEEEauthorblockA{\textit{Department of Computer and Software Engineering, Polytechnique Montreal, Montreal, Canada} \\
\textit{ samar.abdelghani@polymtl.ca, soumaya.cherkaoui@polymtl.ca}}\\
}
\maketitle
\thispagestyle{empty}
\pagestyle{empty}
\begin{abstract}
Organizations and enterprises across domains such as healthcare, finance, and scientific research are increasingly required to extract collective intelligence from distributed, siloed datasets while adhering to strict privacy, regulatory, and sovereignty requirements. Federated Learning (FL) enables collaborative model building without sharing sensitive raw data, but faces growing challenges posed by statistical heterogeneity, system diversity, and the computational burden from complex models. This study examines the potential of quantum-assisted federated learning, which could cut the number of parameters in classical models by polylogarithmic factors and thus lessen training overhead. Accordingly, we introduce {\sysName}, a quantum-enabled federated learning framework aimed at boosting computational efficiency across edge device networks. We evaluate the proposed framework using the widely adopted FashionMNIST dataset. Experimental results show that {\sysName} achieves a 77.6\% reduction in the parameter count of a VGG-like model while maintaining an accuracy comparable to classical approaches in a scalable environment. These results point to the potential of leveraging quantum computing within a federated learning context to strengthen FL capabilities of edge devices.

\end{abstract}

\IEEEpeerreviewmaketitle
\section{Introduction}
The increasing demand for data-driven intelligence across diverse sectors, such as healthcare, finance, autonomous systems, and scientific research, is often at odds with stringent privacy, security, and regulatory constraints. Many organizations today are unable to centralize or share sensitive raw data, yet the collective learning from distributed, siloed datasets remains essential to deliver robust, generalizable ML models~\cite{9220170, abouaomar2022federated}. 

Federated learning (FL)~\cite{9524974} has emerged as a transformative approach, allowing multiple institutions to jointly train a global model by synchronizing local parameter updates, while their data remains strictly on-premise~\cite{GDPRR}. These collaborative protocols not only preserve data locality and sovereignty, but also mitigate communication overhead and reduce compliance barriers. However, as FL adoption widens, new technical bottlenecks arise: heterogeneous client hardware, statistical heterogeneity (non-IID data), and—critically—the computational and communication overload imposed by modern deep models~\cite{taik2022clustered,taik2021data_aware,fedpylot}. While compression and pruning ~\cite{quantization_mixed} can reduce model size, reduce expressivity or model accuracy.\\
Quantum machine learning (QML) has recently emerged as an interesting paradigm ~\cite{qml2022challenges, 10.1145/3616388.3625543}. Central to this paradigm are variational quantum circuits (VQCs), which serve as the core QML models. VQCs encode classical data into quantum states, process these states via parameterized quantum gates, and then measure them to obtain classical output. Although they often use far fewer tunable parameters than comparable classical networks, they can converge more quickly in certain settings. VQCs maintain strong representational power, making them valuable building blocks in QML architectures~\cite{quantum-train-idea}. Under certain conditions, VQCs have been theoretically demonstrated to outperform their classical counterparts. In practice, VQC-based models have shown success in a wide range of machine learning tasks, such as time-series analysis, classification, anomaly detection, and other tasks~\cite{10821473,10811384,11059562,KalfonCherkaouiLaprade2024,mlika2023user}. In particular, the Quantum-Train (QT) framework~\cite{quantum-train-idea,trainingML-QML,fed-qt-minst} offers a hybrid quantum–classical training protocol in which a quantum neural network (QNN) generates the parameters of a much larger classical model through a classical mapping function. This architecture has the potential to enable a polylogarithmic reduction in trainable parameters, improving efficiency while often preserving or enhancing learning capacity. Crucially, after training, the resulting model remains fully classical, eliminating  quantum hardware for inference.

\begin{figure*}[t]
    \centering    \includegraphics[width=0.99\textwidth]{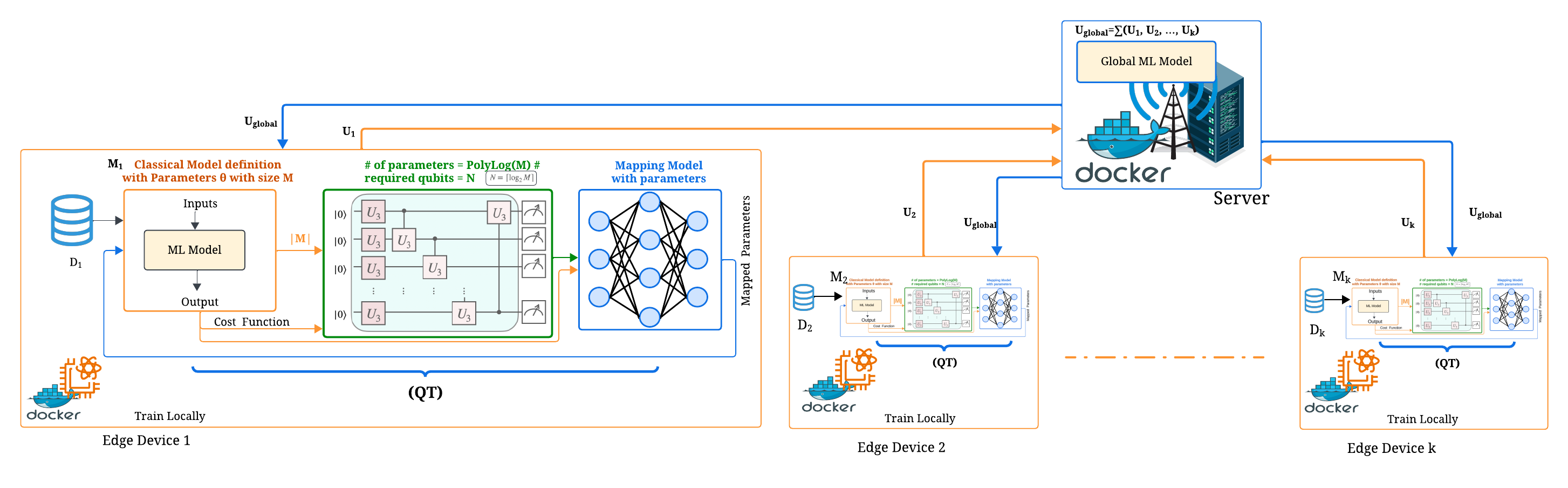}
    \vspace{-3mm}
    \caption{The figure illustrates three key aspects: (1) the overall FL workflow, showing the interaction between edge devices and the central server; (2) a detailed view of one edge device with its core components: classical ML model, QNN, and mapping model; and (3) the proposed federated architecture that integrates QT, with edge devices simulated using Docker containers.}  
    \label{fig:FL-workflow-qt-system}
   \vspace{-4mm}
\end{figure*}

In this work, we introduce QFed: a quantum-enabled federated learning framework that embeds the QT approach into local training at each participant. QFed enables each participant to independently leverage quantum-enhanced model compression, with the objective of reducing communication and computational overhead by exchanging only compact classical model updates within the federation.  Participants do not require on‑device quantum processors; they access cloud‑hosted quantum hardware or run quantum‑circuit simulators on classical machines, aligning with current NISQ realities. This approach reflects the realities of current quantum technology, where direct integration of quantum processors in edge devices is not yet practical or widespread. Instead, quantum-enhanced federated learning is made accessible to participants through remote, virtualized, or cloud-hosted quantum computing resources, ensuring feasibility for a broader range of distributed or intelligent edge platforms beyond minimal or strictly constrained IoT devices.

To rigorously assess QFed, performance is benchmarked against conventional federated learning with full‑sized classical models. Experiments on FashionMNIST emulate realistic participant diversity, varying client counts, local epochs, and communication rounds under identical partitioning and aggregation. This setup enables a direct comparison of model compactness, training and communication characteristics, and accuracy with standard, non‑compressed classical FL baselines.
The main contributions of this paper are:
\begin{itemize}
 \item[\textbf{(1)}] \textbf{QFed framework.} A federated learning paradigm that integrates QT approach into client‑side training to generate parameter‑efficient classical updates while keeping inference fully classical.
 \item[\textbf{(2)}] \textbf{Empicrical Evaluation.} A controlled study on FashionMNIST across varied client counts, communication rounds, and local epochs, compared against non‑compressed classical federated baselines under matched training protocols.
 \item[\textbf{(3)}] \textbf{Containerized testbed.} A Docker/MPI environment emulates resource heterogeneity and inter‑client communication to support reproducible experiments and extensions.
\end{itemize}
This work aims to test whether hybrid quantum–classical federated optimization can improve model compactness and resource efficiency in privacy-preserving, distributed learning scenarios, and to explore the real-world practicalities and challenges of such integration.


\section{Preliminaries}
This section presents the foundational concepts underlying our work, including an overview of Federated Learning principles and key quantum computing concepts, with particular emphasis on the Quantum-Train paradigm.
\subsection{Federated Learning}
Federated Learning is a distributed machine learning framework that enables multiple devices to collaboratively train a global model while ensuring that local data remain decentralized. This approach provides several unique advantages. Firstly, by transmitting local model parameters rather than raw user data, FL promotes data privacy and enables the central model to be trained on decentralized data without exposing sensitive information, and also ensures compliance with data protection regulations such as GDPR~\cite{GDPRR}. Secondly, communication overhead is significantly reduced in large-scale systems, since the size of model parameters is typically much smaller than that of the training data.

The general workflow of FL is depicted in Figure~\ref{fig:FL-workflow-qt-system}. As shown in this figure, for each edge device $k$, a local model $M_K$ is trained using the edge’s private dataset $D_K$. The resulting model parameters $U_K$ are sent to a central server, where they are aggregated using an aggregation function, represented as $U_{\text{global}} = \sum (U_1,U_2,...,U_k)$. Here, $\sum(\cdot)$ denotes the aggregation function and $U$ represents the updates from each device. The aggregated parameters $U_{\text{global}}$ are then distributed back to the clients to update their local models, enabling further training in the next round.
\subsection{Quantum Computing and Quantum Train}
\subsubsection{Quantum Computing (QC)}
In classical computing, the bit is the fundamental unit of information that exists only in one of two states: 0 or 1. In contrast, in quantum computing, the basic unit is the qubit, which can exist in a superposition of both states simultaneously, described by the equation:
\[
|\psi\rangle = a|0\rangle + b|1\rangle, \quad \text{where } |a|^2 + |b|^2 = 1.
\]
This probabilistic state is maintained until a measurement is performed, which collapses the qubit into one of the basis states (e.g., $|0\rangle$ or $|1\rangle$). Just as classical computing uses logic gates (e.g., AND, OR, NOT), quantum computing uses quantum gates that operate on qubits. These gates are represented by unitary matrices, which means that they preserve the total probability (i.e., the norm of the qubit state). One of the fundamental quantum gates is the X gate (also known as the quantum NOT gate), which flips the qubit state from $|0\rangle$ to $|1\rangle$ and vice versa~\cite{quantum-terms-qdiff}. Its matrix form is:
\\\\\\
\[
X = \begin{bmatrix} 0 & 1 \\ 1 & 0 \end{bmatrix}.
\]
A more general gate used in quantum algorithms is the \textbf{U3 gate}, a universal single-qubit gate parameterized by three rotation angles $\alpha$, $\phi$, and $\lambda$, with the matrix:
\[
U3(\alpha, \phi, \lambda) = \begin{bmatrix}
\cos(\alpha/2) & -e^{i\lambda} \sin(\alpha/2) \\
e^{i\phi} \sin(\alpha/2) & e^{i(\phi + \lambda)} \cos(\alpha/2)
\end{bmatrix}.
\]
This gate enables arbitrary rotations on the Bloch sphere, which is a geometrical representation of the state space of a single qubit, where each pure quantum state corresponds to a point on the surface of a unit sphere. U3 gate is widely used in VQCs and QNNs. Furthermore, the \textbf{CU3 gate} is the controlled version of U3, applying the U3 operation to a target qubit only when the control qubit is in state $|1\rangle$. CU3 is particularly useful for building parameterized entanglement between qubits, which is crucial for advanced quantum learning architectures such as those in hybrid quantum-classical models.

Like logic circuits in classical computing, a quantum circuit is a sequence of quantum gates whose overall operation is the product of their unitary matrices ~\cite{quantum-terms-qdiff}. For example, the middle green block within the QT-labeled section of Figure~\ref{fig:FL-workflow-qt-system} represents the QNN circuit, where the qubits are initialized in the \(|0\rangle\) state. The circuit employs gates \(U3\) and \(CU3\), and measurements are performed at the end of the circuit~\cite{trainingML-QML}.

Moreover, quantum algorithms are computational procedures designed to operate on quantum hardware, typically requiring multiple repeated executions due to the inherent probabilistic nature of quantum measurement. Notable examples include Shor’s algorithm for factoring, Grover’s algorithm for search, and Variational Quantum Algorithms (VQAs). VQAs form the foundation of many current QML and quantum optimization techniques. QML is an emerging interdisciplinary field at the intersection of quantum computing and classical machine learning. By harnessing fundamental quantum mechanical phenomena such as superposition, interference, and entanglement, QML offers a powerful computational paradigm. This enables potential advantages in tasks such as classification, optimization, and generative modeling~\cite{qml2022challenges}.

\subsubsection{Quantum-Train (QT)}
Unlike traditional QML approaches that rely on quantum models for both training and inference, QT utilizes QNNs solely during the training phase to tune parameters for classical models. Once trained, the classical model operates independently, requiring no quantum resources for inference. This decoupling from quantum systems during inference makes QT particularly attractive and practical, especially given the current limitations of quantum hardware. In Figure~\ref{fig:FL-workflow-qt-system}, within edge device 1—highlighted by the QT label—the process begins by feeding classical inputs into a classical ML model to generate outputs. Based on the number of parameters in this model, the corresponding number of qubits needed for the QNN is determined. The QNN is then executed to adjust its internal parameters using the cost function of the classical model as guidance. In the final stage, the quantum measurement results are mapped back to update the parameters of the classical model. This hybrid training loop is repeated iteratively until the model achieves convergence. Further details are in~\cite{quantum-train-idea,trainingML-QML}.


\section{Related Works}
Recent advances in quantum computing QML have created new possibilities for enhancing FL frameworks. Quantum federated learning (QFL) combines quantum computational tools with FL as highlighted in recent surveys and position papers~\cite{qiao2024transitioning,kwak2023qddlmodelsdiscussion}. Early work on QFL focused on using QML as local and/or global learners, typically requiring quantum hardware during training and inference~\cite{xia2021quantumfed,chen2021federatedQML}. However, reliance on quantum end-devices limits practical deployment in current settings~\cite{qiao2024transitioning,pira2023invitation}.\\ 
Classical FL has adopted a range of compression methods (e.g., pruning, quantization) to reduce parameter count and hence communication cost~\cite{9762360,9054168}. Similarly, some works have proposed the concept of quantum neural network compression~\cite{hu2022quantum}.
In parallel, efforts have sought to address practical deployment challenges through hybrid approaches. The QT framework~\cite{quantum-train-idea,trainingML-QML,fed-qt-minst} employs QNNs exclusively during the training phase to generate parameters for classical models. This approach minimizes the number of trainable parameters and eliminates the need for quantum inference hardware~\cite{fed-qt-minst} to circumvent hardware constraints typical of NISQ-era and cater to distributed and federated scenarios. Empirical studies are currently scarce, with only a few integrating QT into federated settings and testing at scale with heterogeneous clients~\cite{fed-qt-minst,IoTstreaming}. Other QFL efforts are mostly theoretical or consist of proof-of-concept implementations that do not scale to real-world scenarios or account for system diversity~\cite{kim2023quantum-AV}.

QFed extends the landscape by providing a containerized, scalable system for quantum-assisted federated learning. By confining quantum computation to the training phase and maintaining classical inference, this work offers a pragmatic step toward harnessing quantum advantages within the constraints and requirements of real-world federated systems.
\section{System Model}

The proposed {\sysName} workflow is illustrated in Figure~\ref{fig:FL-workflow-qt-system}. As shown, the edge devices are simulated using Docker containers to reflect their  computational resources by adjusting their capabilities. Communication between edge devices and server is emulated by constructing an MPI (Message Passing Interface) cluster of containers, facilitating seamless deployment of the proposed {\sysName} framework. Each edge device/container is assumed to have access to a Quantum Processing Unit (QPU). 
Following the standard FL pipeline, each container performs local training, where model parameters are optimized and trained using quantum computation. Due to the limitations of the NISQ era, these computations are executed on quantum simulators. The proposed pipeline is broken down into the following steps.

\begin{enumerate}
    \item \textbf{Target Classical Network:} A classical ML model with a parameter vector $\theta$ of size $M$ is selected, corresponding to the left orange block within edge device 1 in Figure~\ref{fig:FL-workflow-qt-system}.
    
    \item \textbf{Quantum Neural Network (QNN):} A QNN with $N = \lceil \log_2 M \rceil$ qubits and parameter vector $\beta$ is constructed using a variational quantum circuit (e.g., built with gates such as $U_3$ and $CU_3$), as represented in the middle green block within edge device 1 in Figure~\ref{fig:FL-workflow-qt-system}.
    
    \item \textbf{Measurement:} The quantum state $|\psi(\theta)\rangle$ is measured in the computational basis, producing $2^N$ probabilities of the form $|\langle\phi_i|\psi(\theta)\rangle|^2$, corresponding to the last step in the middle green block of Figure~\ref{fig:FL-workflow-qt-system}.
    
    \item \textbf{Classical Mapping Model:} These probabilities are fed into a classical mapping model (typically an MLP model with parameters $\gamma$) to generate the classical network’s parameters $\theta$. The \textbf{tanh} function is used here, inspired by the tanh activation function~\cite{trainingML-QML}, as shown in the right blue block within edge device 1 in Figure~\ref{fig:FL-workflow-qt-system}.
    
    \item \textbf{Local Training:} The classical model is trained on labeled data,  gradients backpropagated via classical and quantum layers to update parameters $\theta$, $\beta$, and $\gamma$.
    
    \item \textbf{Sending Model Parameters:} After local training, containers send updated parameters to a central server, which aggregates them to update the global model.

    \item \textbf{Building the Global Model:} At the server side, the local model parameters from all containers are received, the server aggregates them to build an updated global model, which is then sent back to the edge devices/containers.
    
    \item \textbf{Achieving Convergence:} This iterative process—local training, quantum-based tuning at each container, followed by sending the model parameters to the server and receiving the updated global model—continues until convergence is achieved.
    
    \item \textbf{Inference:} Finally, after training, only the classical model is used for inference, completely removing the need for quantum computations in the deployment phase.
\end{enumerate}
\section{Results and Analysis}

\begin{figure}[htbp]
    \centering
    \includegraphics[width=0.45\textwidth]{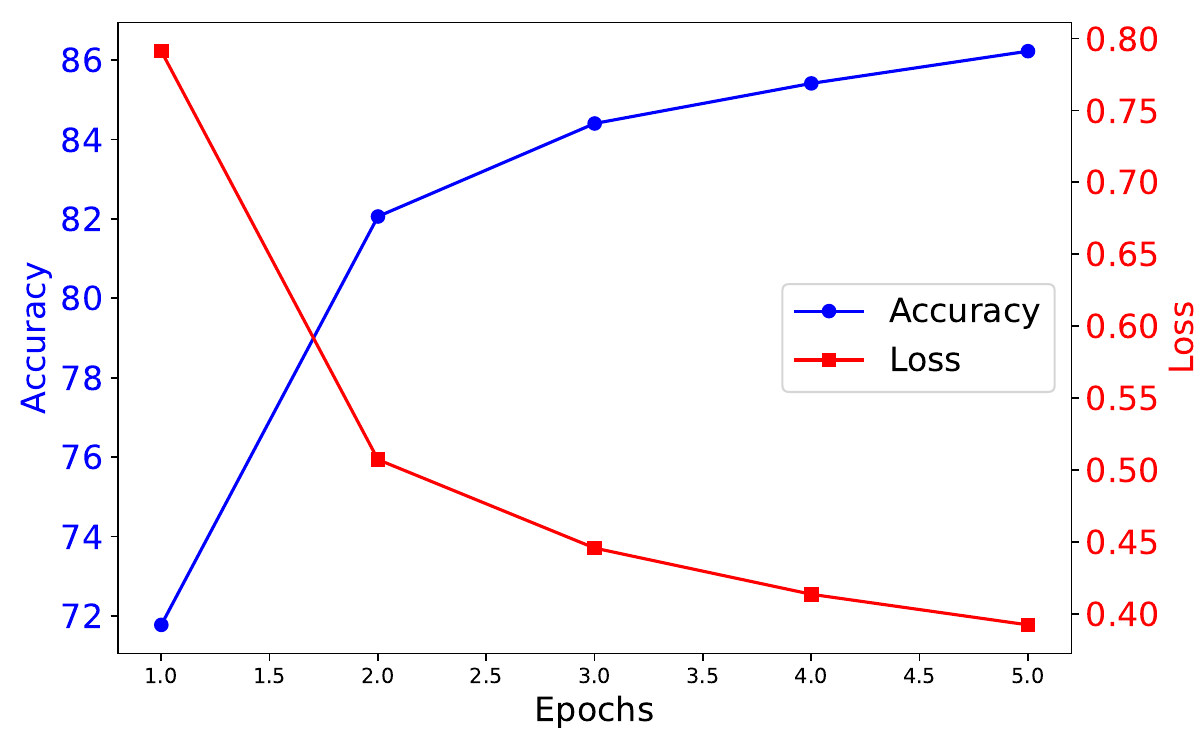}
    \vspace{-2mm}
    \caption{Accuracy and loss curves for centralized classical training without Quantum-Train.}  
    \label{fig:Cmodel-centralized}
    \vspace{-1em}
\end{figure}

All experiments were conducted on a system running Ubuntu 24.04, equipped with an Intel Core i9-13900HX processor (32 GB RAM) and an NVIDIA RTX 4070 GPU. A VGG-like classical model was applied to the Fashion-MNIST dataset~\cite{fashion-MNIST} for training and evaluation. The training was carried out on quantum simulators deployed within isolated container environments. The source code is available at \href{https://github.com/SamarShabanCS/QFed-Parameter-Compact-Quantum-Classical-Federated-Learning}{GitHub}. 
In a centralized, purely classical computing setting, the model achieved an accuracy of 86.09\% with a total of 6,690 trainable parameters. Larger and more complex convolutional neural networks (CNNs) were not employed due to the barren plateau phenomenon, which can lead to vanishing gradients during quantum model training. Figure~\ref{fig:Cmodel-centralized} shows the model’s accuracy and loss over training epochs. As illustrated, performance improves steadily with continued training, suggesting that the proposed classical architecture is capable of effectively learning meaningful representations from the data.
\begin{figure}[htbp]
    \raggedleft
    \includegraphics[width=0.48\textwidth]{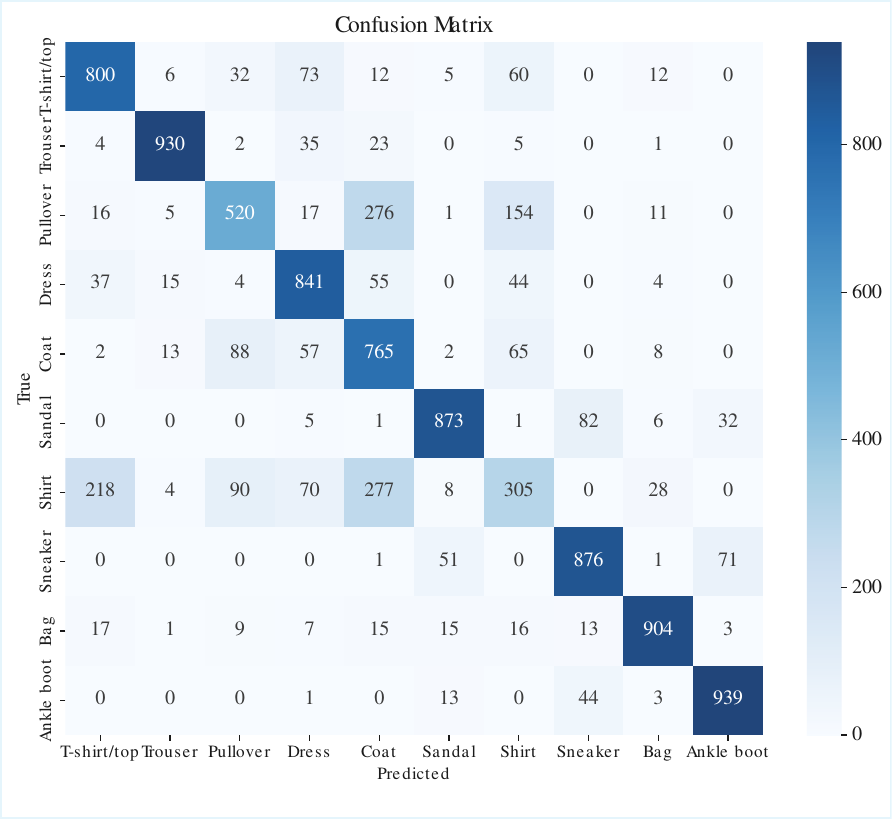}
    \vspace{-3mm}
    \caption{The confusion matrix of the centralized model trained using the Quantum-Train approach.}  
    \label{fig:C_qmodel-cf}
   \vspace{-5mm}   
\end{figure}

\begin{figure*}[htbp] 
    \centering
    \begin{subfigure}[b]{0.45\textwidth} 
        \includegraphics[width=\linewidth]{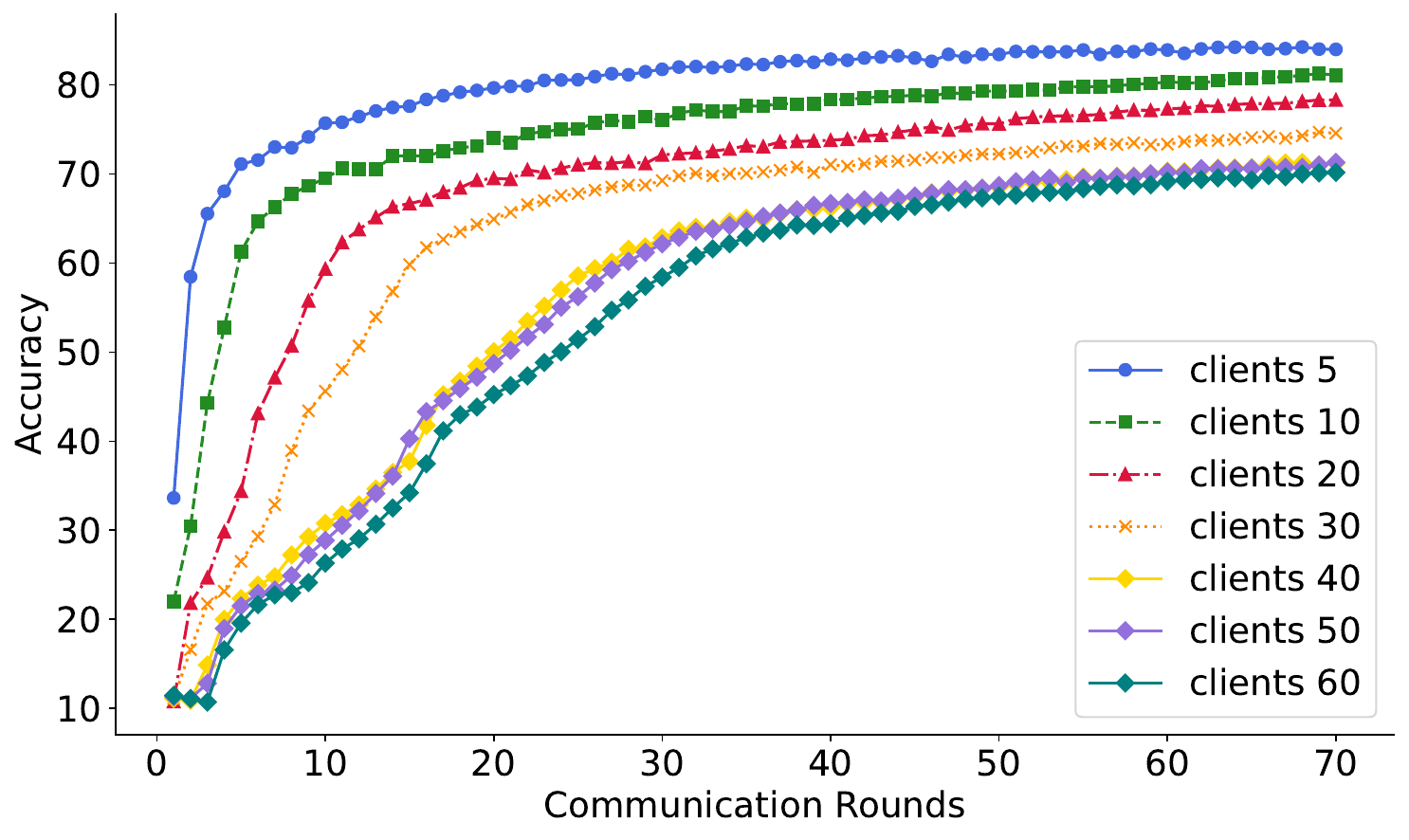}
        \caption{Global Accuracy}
        \label{fig:acc}
    \end{subfigure}
    \hfill 
    \begin{subfigure}[b]{0.45\textwidth}
        \includegraphics[width=\linewidth]{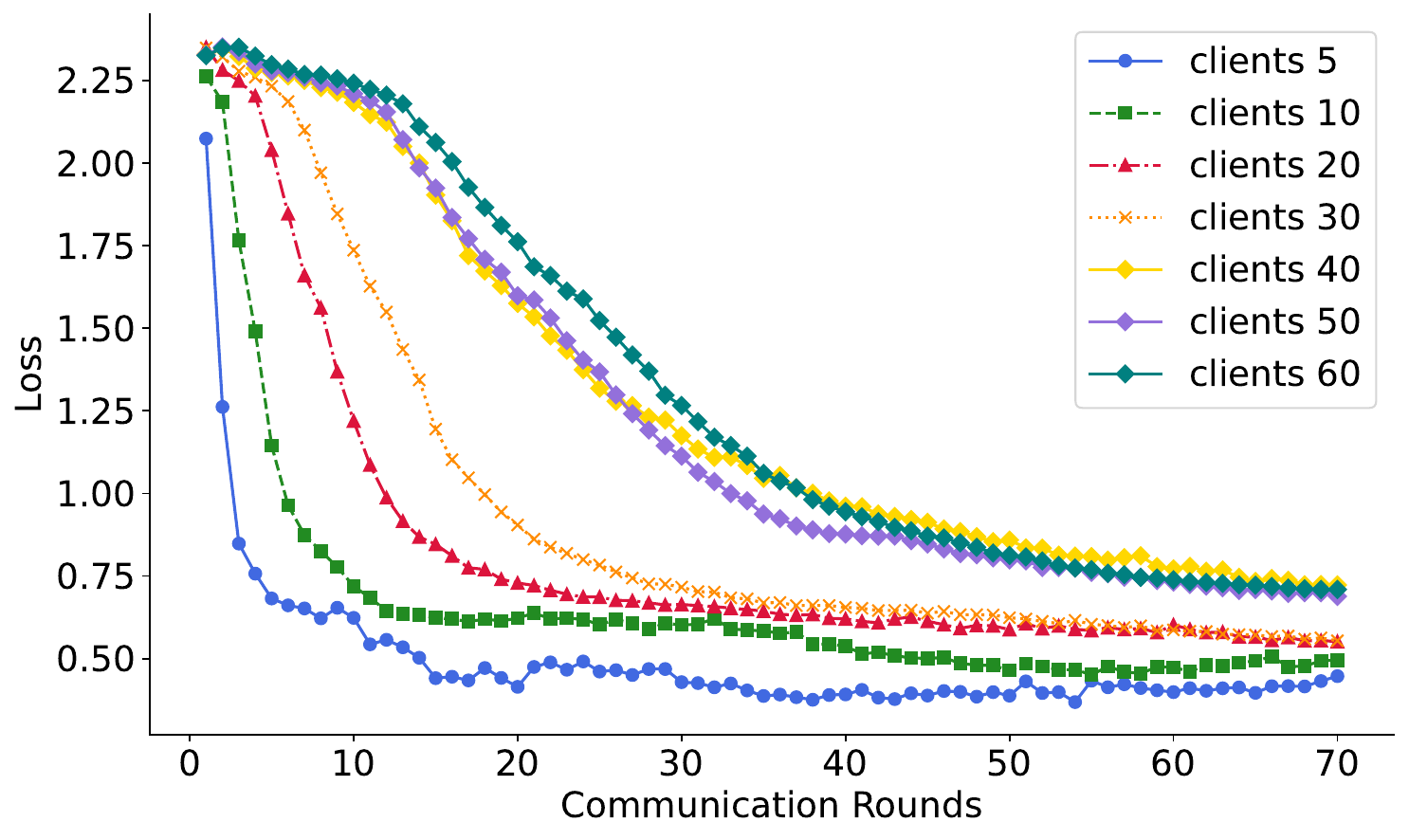}
        \caption{Global Loss}
        \label{fig:loss}
    \end{subfigure}
    \caption{Federated global model performance using QT, accuracy and loss across different clients while communication rounds=70 and local epochs=10, number of clients/edges are tested with 5, 10, 20, 30, 40, 50, and 60.}
    \vspace{-3mm}
    \label{fig:flqt-acc-loss-scale}
\end{figure*}

\begin{figure}[htbp]
    \centering
    \includegraphics[width=0.48\textwidth]{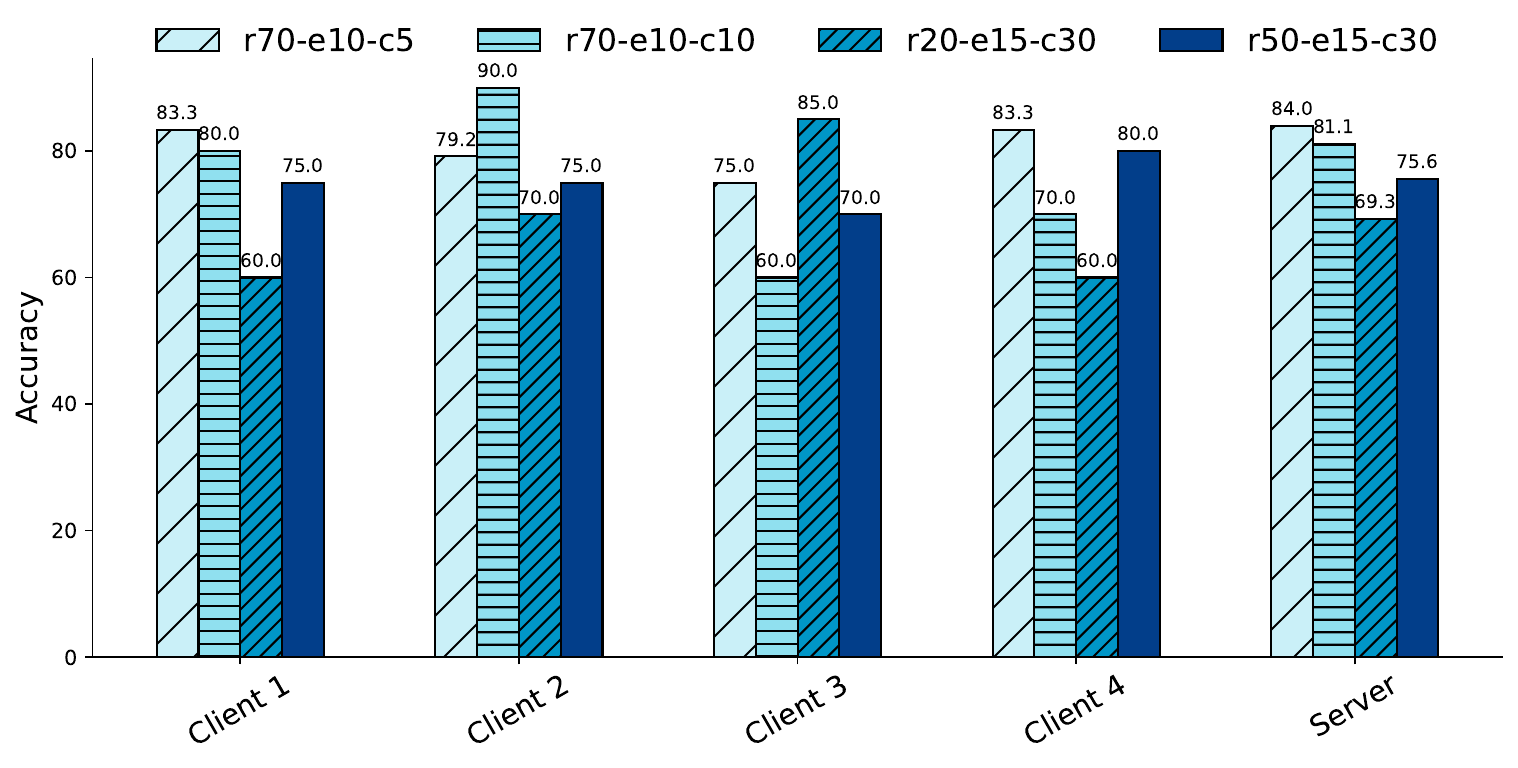}
    \caption{The global and local models' accuracies with different training settings, showing only the first 4 clients and the server.}  
    \label{fig:flqt-convergence}
   \vspace{-5mm}
\end{figure}

To facilitate the transition into quantum computing, we initially evaluated the model in a centralized quantum setting to assess its training capabilities. The {\sysName} system was implemented using TorchQuantum, comprising 16 quantum neural network (QNN) layers with 13 qubits. In this configuration, the quantum model achieved a test accuracy of 78\%.

To evaluate class-wise performance and assess the extent to which quantum training preserves the behavior of the classical model, we analyzed the confusion matrix of the quantum-trained model in the centralized setting (Figure~\ref{fig:C_qmodel-cf}). The analysis revealed notable misclassifications between the "shirt" and "pullover" classes. These results suggest that the inclusion of additional representative examples from these categories may contribute to improved classification performance.

Figure~\ref{fig:flqt-acc-loss-scale} illustrates the performance of the FL-QT approach, indicating a consistent increase in model accuracy accompanied by a steady decline in loss. While the accuracy of the QT-FL model decreases with a higher number of participating clients, the convergence of the loss curves suggests that further training epochs may be necessary to achieve generalization. These observations indicate that the model is capable of scaling to larger federated settings. 
The observed robustness is attributed to the aggregation of richer local updates, which capture the heterogeneity of client data distributions and contribute to efficient and generalized data distribution learning.

Figure~\ref{fig:flqt-convergence} provides further insight into the efficiency of our federated learning framework by evaluating its generalization behavior. The results of local and global model performance across different experimental configurations vary in the number of: local training epochs, communication rounds, and participating clients. For example, the r70-e10-c10 experiment refers to 70 communication rounds, 10 local epochs, and 10 participating clients. The figure shows results for only the first four clients per setting. These diverse configurations demonstrate the flexibility of the {\sysName} approach in real-world scenarios, where client data heterogeneity leads to differences in local model accuracies but does not adversely affect the generalization of the global model. This is particularly evident in the r20-e15-c30 setting, which shows consistent performance across various conditions. Regarding parameter reduction, our results indicate a 77.6\% decrease in model complexity when using FL with QT compared to the classical approach, reducing the number of parameters from 6,690 to 1,497. Despite this substantial compression, FL with QT maintains competitive accuracy, achieving 84\% compared to the classical model’s 86\% in centralized training.

\section{Discussion}
Despite the promising results achieved with QT framework and and its integration into federated learning via QFed, several important limitations remain inherent to the current state of quantum technology.
First, hardware constraints in the NISQ era persist: available quantum devices are characterized by limited qubit counts, short coherence times, and significant noise. These factors set practical upper bounds on the size and depth of quantum circuits, thus restricting the complexity of quantum models that can be effectively trained. Consequently, the scale of classical models that QT can meaningfully compress is limited, and the method does not yet scale to the largest neural architectures commonly deployed in industry.\\ Second, expressivity and training challenges arise in variational quantum circuits. QT relies on optimizing variational parameters, but such circuits are susceptible to barren plateaus—regions in parameter space where gradients vanish and training may stagnate. The expressive power of quantum models is fundamentally bounded by the available quantum resources and chosen architectures, which can restrict performance gains compared to high-capacity classical networks.\\
Third, noise and measurement issues remain a challenge. Although QT confines quantum computation to the training phase, the accuracy and reliability of parameter updates remain sensitive to quantum noise and errors. This is particularly true on real NISQ hardware, where measurement accuracy and error mitigation limit the fidelity of trained models.\\
Finally, while QFed demonstrates substantial parameter reductions and computational savings for moderately sized networks, further upscaling to very deep or wide architectures is not yet possible given the aforementioned limitations on model size, circuit connectivity, and error rates.\\
Nonetheless, ongoing advances such as improved quantum error correction, hardware innovations like topological qubits, and more efficient circuit design, raise optimism that these constraints will be alleviated in the near future. As quantum devices become more robust and scalable, QT-based compression for federated learning could be able to handle larger models, achieving greater compression and efficiency benefits. While practical barriers remain, the hybrid quantum-classical paradigm of QT and QFed suggests a potential pathway for quantum acceleration in federated learning.
\section{Conclusion}

Our experiments demonstrate that {\sysName} achieves a 77.6\% reduction in parameter count, enabling efficient operation even on limited hardware and facilitating scalable deployment across diverse edge and organizational networks. Critically, the resulting model remains fully classical, supporting straightforward deployment on existing devices without the need for quantum computing infrastructure during inference. Throughout all evaluated FL scenarios using the FashionMNIST dataset, {\sysName} maintained accuracy comparable to classical approaches despite substantially reducing model complexity.\\
This work is situated in the NISQ era, where quantum hardware remains limited by qubit counts and noise. These constraints limit the size of quantum circuits, and thus the classical models that QT (and subsequently {\sysName}) can effectively compress. Nevertheless, QT’s hybrid scheme allows substantial compression by leveraging a small quantum core to generate parameters for a larger classical model, allowing polylogarithmic parameter reductions without critical loss of expressivity. {\sysName} extends these benefits to realistic federated settings with compressed VGG-Style architectures.\\
Looking ahead, ongoing developments in quantum technology, including advances in error mitigation, error correction, and emerging quantum architectures such as those promised by topological qubits, are expected to raise the ceiling on circuit and model size. As quantum devices improve, even greater model compression and efficiency gains are likely to become attainable in practical deployments. Since {\sysName}’s design strictly confines quantum computation to the training phase, while keeping inference fully classical, it is well-positioned to harness these advances without introducing burdens for real-world deployment. With continuing progress, QFed and related approaches may play an increasingly central role in efficient, privacy-preserving federated machine learning at scale.
\ifCLASSOPTIONcaptionsoff
  \newpage
\fi
\bibliographystyle{IEEEtran}
\bibliography{IEEEabrv,ref}
\end{document}